\documentclass[final]{article}

\usepackage[utf8]{inputenc}
\usepackage[T1]{fontenc}
\usepackage{hyperref}
\usepackage{url}
\usepackage{booktabs}
\usepackage{amsfonts}
\usepackage{nicefrac}
\usepackage{microtype}
\usepackage{graphicx}
\DeclareGraphicsExtensions{.pdf}
\graphicspath{ {/home/jales/report_ntm} }
 
\title{Neural Turing Machines: \\ Convergence of Copy Tasks}
\author{Janez Ale\v{s} \\ {\it Heidelberg Colaboratory for Image Processing} \\
{\it IWR, Heidelberg University} \\ {\it 69120 Heidelberg, Germany }\\ {\small\it janez.ales@iwr.uni-heidelberg.de}}
\date{\today}

\begin{document}
\maketitle
\begin{abstract}
The architecture of neural Turing machines is differentiable end to end and is trainable with gradient descent methods. 
Due to their large unfolded depth Neural Turing Machines are hard to train and 
because of their linear access of complete memory they do not scale.
Other architectures have been studied to overcome these difficulties.
In this report we focus on improving the quality of prediction of the original linear memory architecture on copy and repeat copy tasks.
Copy task predictions on sequences of length six times larger than those the neural Turing machine was trained on prove to be highly accurate
and so do predictions of repeat copy tasks for sequences with twice the repetition number and twice the sequence length neural Turing machine was trained on.
\end{abstract}
\section{Introduction}
Neural network capabilities were extended by coupling them to linear external memory accessed by probability distribution weight vectors, 
called Neural Turing Machines (NTM) \cite{graves}.
This architecture is differentiable end to end and can be trained with gradient descent.
It is known that their large unfolded depth makes them hard to train.
They also do not scale well due to the linear access of their complete memory.
Alternate memory architectures have since been proposed to improve training behaviour.
Structured memory components for neural Turing machines were tested for speed of convergence and quality 
of predictions in \cite{zhang}.
In contrast to the original architecture neural GPUs are higly parallelizable and scale well~\cite{kaiser}.
\begin{figure}[h]
\label{fi:1}
\includegraphics[width=\linewidth]{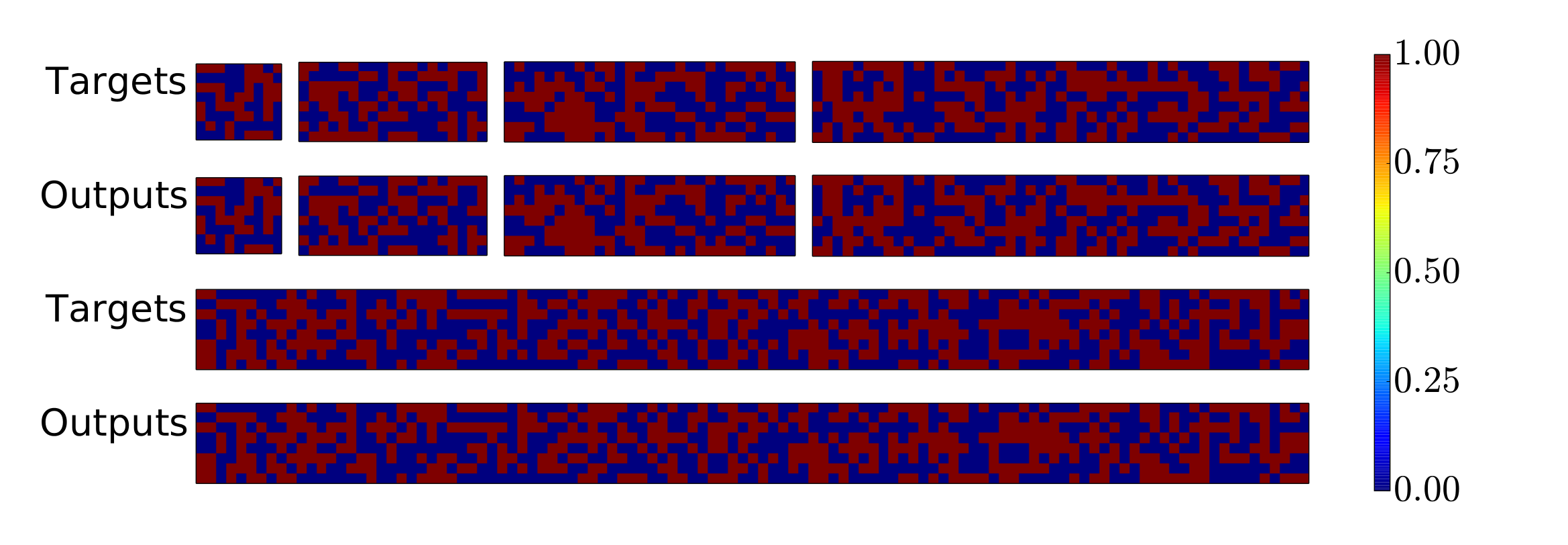}
\hspace*{1.2cm}\includegraphics[width=0.9\linewidth]{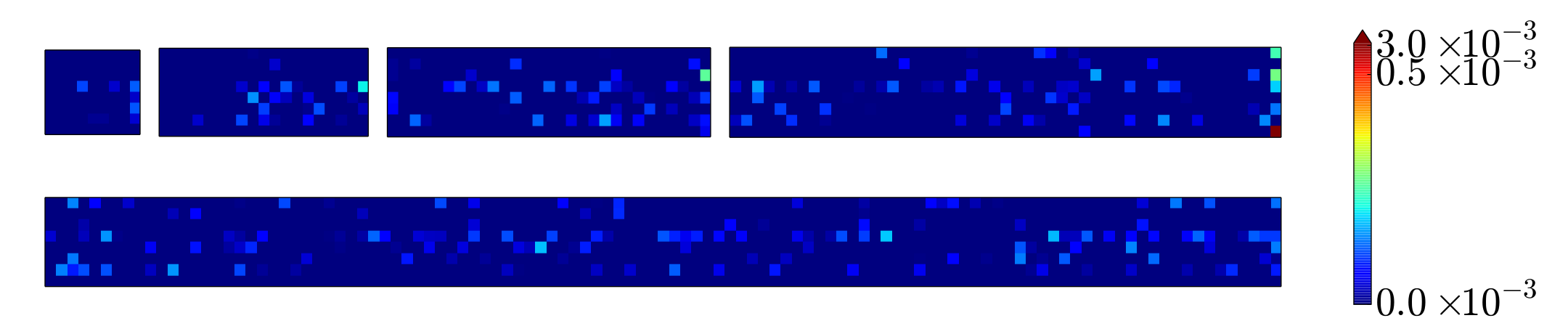}
\caption{Targets and outputs for sequences of length $10$, $20$, $30$, $50$, and $120$, followed by
differences of outputs and targets with a logarithmic colormap.}
\end{figure}

In this report we focus on the quality of predictions of the original linear memory architecture of NTMs (see \cite{graves}) 
for copy and repeat copy tasks on
{\it longer} sequences than those the neural turing machine was trained on. 
Prediction on these sequences prove to be accurate.

\section{Copy Task}
An NTM with a single layer feedforward controler of size $100$ and external memory size $128 \times 20$
is trained for a copy task on sequences of random binary vectors of length $8$ 
with sequence lengths chosen randomly between $1$ and $20$. 
Prediction quality for this task was already examined in \cite{graves}.
Tests on sequences of legth $10$ and $20$ reproduced inputs with high confidence and with virtually no mistakes.
However, predictions on sequences of length $30$ and $50$ show a few errors on a relatively small number of bits.
Predictions on sequences of length $120$ produce local and 
global errors leading to high loss.
A higher accuracy on the copy task than in the original paper was achieved in \cite{rasmus} using structured memory NTMs.

\begin{figure}[h]
\includegraphics[width=0.9\linewidth]{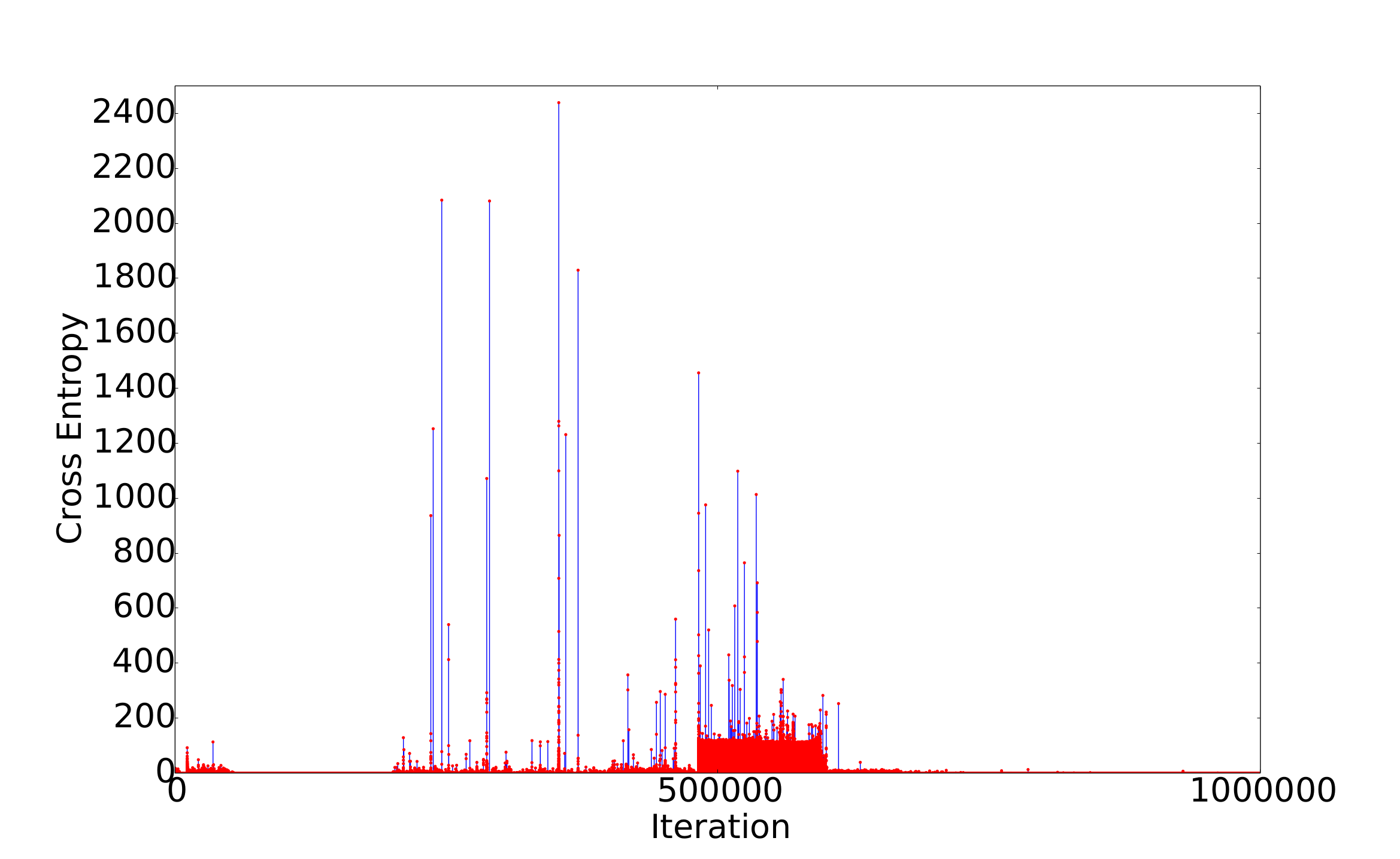}
\caption{Copy learning curve.}
\label{fi:curve}
\end{figure}

We present a solution using the original NTM architecture.
\begin{figure}[t]
\label{fi:2.0}
\includegraphics[width=1.0\linewidth]{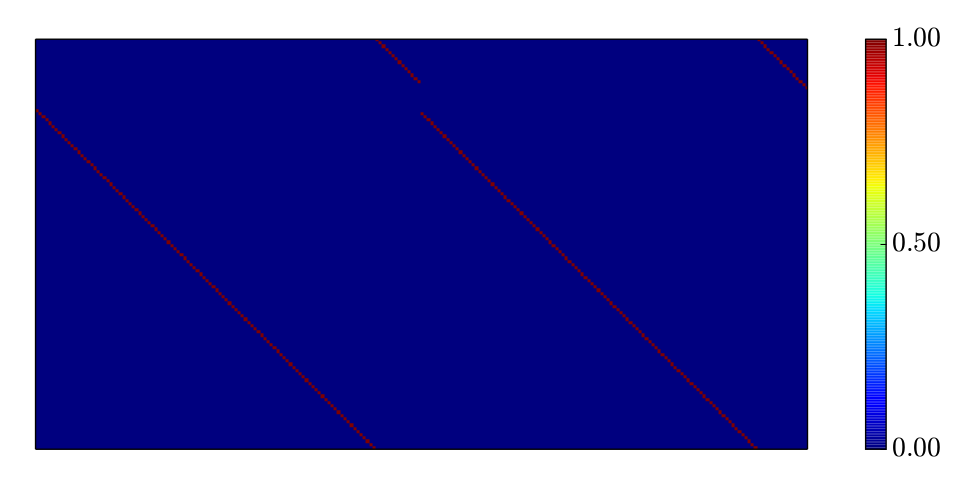}
\caption{Write and read weightings for a sequence of length $120$.}
\end{figure}
Learning stage was carried out on sequences of random length up to $20$
with the sum of bit values {\it not} equal to half of the number of bits. 
Testing was done on the complement set, that is, on the sequences 
with the sum of bit values {\it equal} to half of the number of bits
assuring independent test results.
Predictions reproduced inputs with high confidence for sequences of length $10$, $20$, $30$, $50$, and $120$.
Prediction statistics were calculated on the basis of $10000$ random instances for each sequence length.
We discuss the results of a representative set.
Predictions with no bit errors are shown in Figure~\ref{fi:1}.
\begin{table}[h]
\caption{Prediction statistics for a copy task on a representative set of $10000$ sequences.}
\label{ta:copy}
\begin{center}
\begin{tabular}{|l|r|r|r|r|r|}
\hline
{Sequence length} & {10} & {20} & {30} & {50} & {120} \\
\hline
{Number of seq. with bit errors} & {0} & {0} & {0} & {13} & {36} \\
{Maximum bit error}              & {0} & {0} & {0} & {1} & {1} \\
{Mean of bit errors}             & {0.0000} & {0.0000} & {0.0000} & {0.0013} & {0.0036} \\
{Standard deviation of bit errors} & {0.0000} & {0.0000} & {0.0000} & {0.0360} & {0.0599} \\
\hline
\end{tabular}
\end{center}
\end{table}
Sequence lengths, 
number of sequences with at least one bit error,
maximum bit error,
mean of bit errors, and 
standard deviation of bit errors 
are shown in the Table~\ref{ta:copy}.
The maximum bit error was $1$ and there were no sequences with a global error.
Learning curve for the copy task is shown in Figure~\ref{fi:curve}.

\section{Repeat Copy Task}
\begin{figure}[htb]
\label{fi:3}
\includegraphics[width=0.975\linewidth]{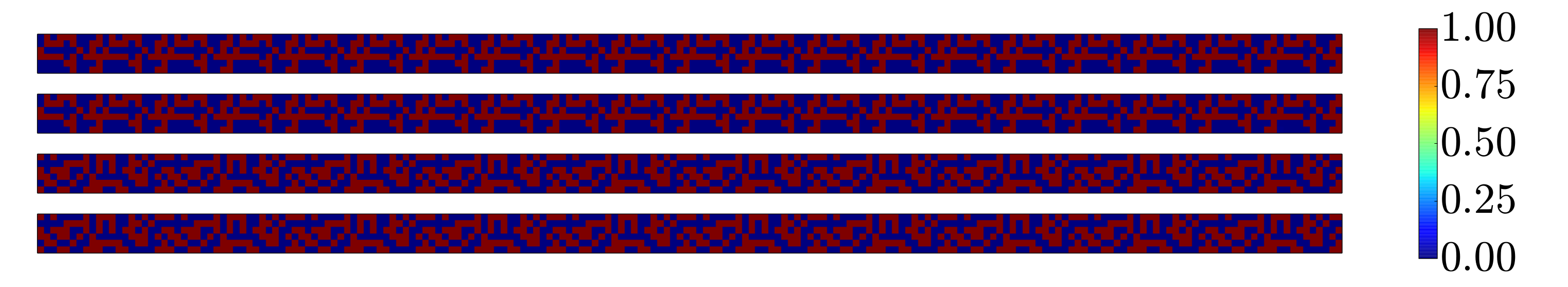}
\includegraphics[width=1.0\linewidth]{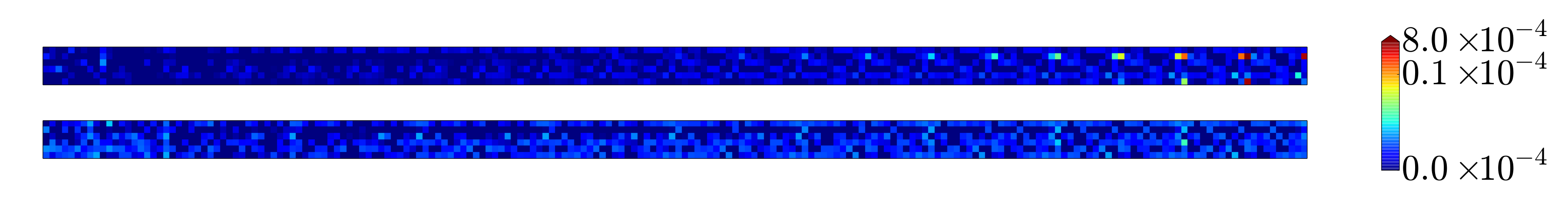}
\caption{Targets and outputs of the repeat copy task for sequences of length $10$ with repetition number $20$ and length $20$ with repetition number $10$,
followed by the corresponding differences.}
\end{figure}
As in the original paper we train the NTM on $6$ bit vector sequences of length up to $10$ and repetition number up to $10$.
We test the quality of the prediction on sequences of length $10$ with repetition number $20$,
on sequences of length $20$ with repetition number $10$, and
on sequences of length $20$ with repetition number $20$.
Prediction statistics was based on $10000$ random sequences.
We discuss a representative result.
\footnote{Predictor weights can be provided upon request.}

Prediction for sequences of length $20$ with repeat number $10$ had
$78$ sequences having at least one faulty bit.
Maximum bit error was $7$ with bit error mean of $0.013$ and bit error standard deviation of $0.183$.

Prediction for sequences of length $20$ with repeat number $20$ had
$104$ sequences having at least one faulty bit.
Maximum bit error was $10$ with bit error mean $0.018$ and bit error standard deviation of $0.254$.

Prediction on sequences of length $10$ with repeat number $20$ were less accurate with
bit error mean $2.449$ and bit error standard deviation of $8.391$.
In total $4418$ sequences had at least one bit error and
$19$ sequences had more than $100$ faulty bits.

Alternate NTM training sessions of repeat copy task will be addressed in future work.

\begin{figure}[ht]
\label{fi4}
\includegraphics[width=1.0\linewidth]{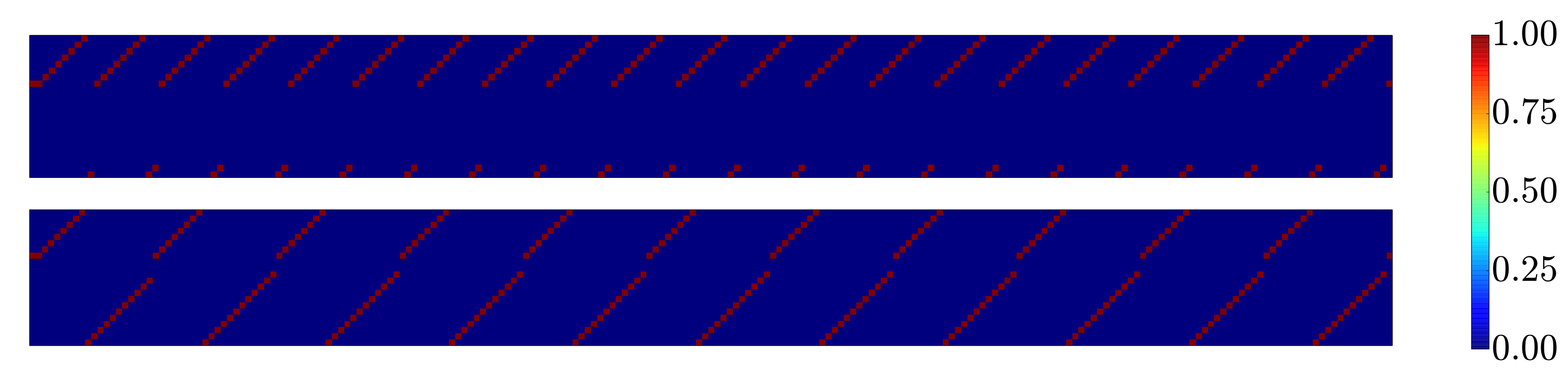}
\caption{Write and read weightings for the repeat copy task for sequences of 
length $10$ with repetition number $20$ and 
length $20$ with repetition number $10$.}
\end{figure}
\begin{figure}[h]
\label{fi:13}
\includegraphics[width=1.0\linewidth]{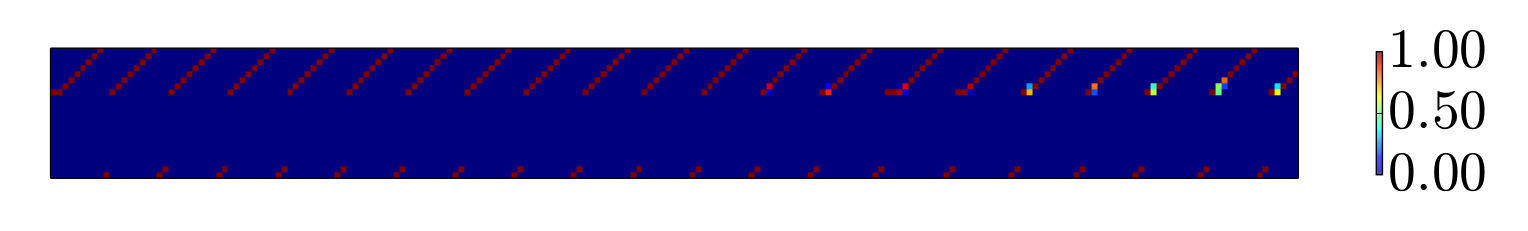}
\caption{Write and read weightings for the repeat copy task for a sequence of length $10$ and repetition number $20$
with a global error.}
\end{figure}

\begin{figure}[h]
\label{fi:11}
\includegraphics[width=0.88\linewidth]{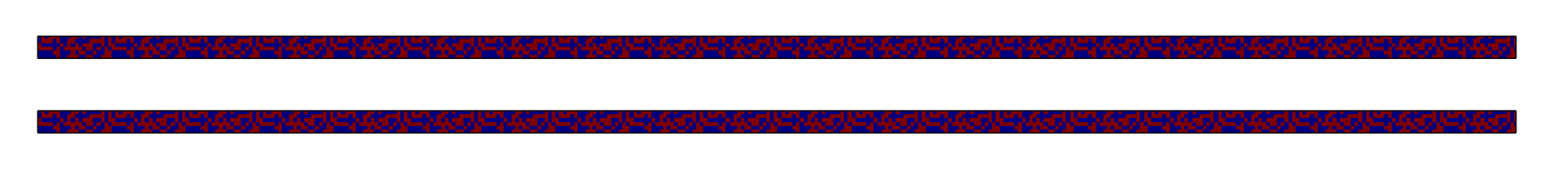}
\includegraphics[width=1.0\linewidth]{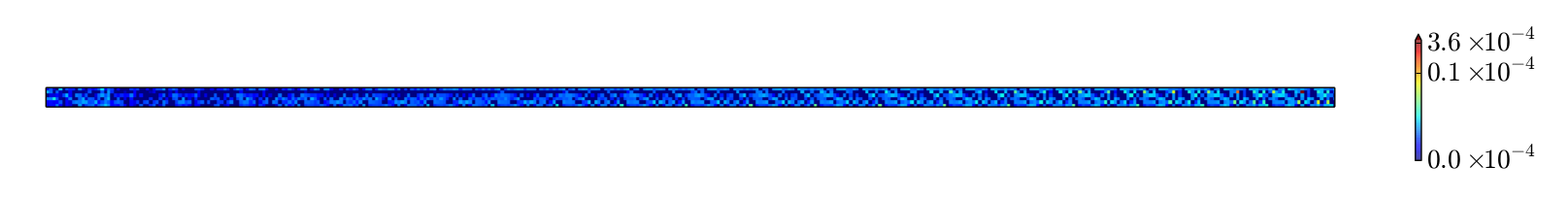}
\includegraphics[width=0.88\linewidth]{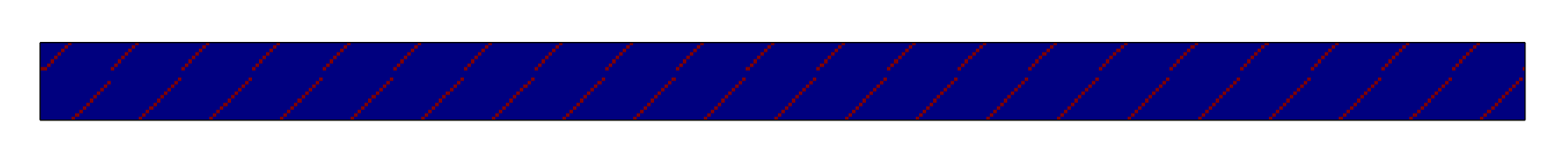}
\caption{Targets and outputs of the repeat copy task for sequences of length $20$ and repetition number $20$.
The corresponding differences are shown in logarithmic scale,
followed by write and read weightings.}
\end{figure}

\pagebreak
\bibliographystyle{plain}
\bibliography{ntm_copy_repeatcopy_tech_report}
\end{document}